\documentclass[runningheads]{llncs}

\usepackage{graphicx}

\usepackage{amsmath,amssymb}

\usepackage{booktabs}
\usepackage{relsize}
\usepackage[export]{adjustbox}
\usepackage{subfigure}
\usepackage{multirow}
\usepackage{makecell}
\usepackage{caption}
\usepackage[dvipsnames]{xcolor}
\usepackage{float}
\newfloat{floatbox}{t}{lob}[section]

\usepackage{amssymb}
\usepackage{pifont}
\newcommand{\cmark}{\ding{51}}%
\newcommand{\xmark}{\ding{55}}%

\begin{document}
\title{Forecasting Future Instance Segmentation with Learned Optical Flow and Warping}
\titlerunning{Forecasting Future Instance Segmentation}
\author{Andrea Ciamarra \and
Federico Becattini \and
Lorenzo Seidenari \and
Alberto Del Bimbo}
\authorrunning{A. Ciamarra et al.}
\institute{University of Florence, Italy \\
\email{\{name.lastname\}@unifi.it}}
\maketitle
\begin{abstract}
For an autonomous vehicle it is essential to observe the ongoing dynamics of a scene and consequently predict imminent future scenarios to ensure safety to itself and others. This can be done using different sensors and modalities.
In this paper we investigate the usage of optical flow for predicting future semantic segmentations. To do so we propose a model that forecasts flow fields autoregressively. Such predictions are then used to guide the inference of a learned warping function that moves instance segmentations on to future frames.
Results on the Cityscapes dataset demonstrate the effectiveness of optical-flow methods.
\end{abstract}
\section{Introduction}
Understanding urban environments from an ego-vehicle perspective is crucial for safe navigation.
This problem can be declined under several points of view, such as detecting entities~\cite{he2017mask}\cite{perot2020learning}, understanding the road layout~\cite{berlincioni2019road} or predicting the future~\cite{marchetti2020multiple}\cite{luc2018predicting}\cite{hu2021apanet}.
At the same time, different modalities can be exploited to understand the environment ranging from RGB, which can produce useful representations such as semantic segmentations~\cite{lin2021predictive}, to depth or even data produced by more complex sensors such as LiDARs~\cite{beltran2018birdnet}, event cameras~\cite{perot2020learning} or thermal cameras~\cite{kieu2020task}.
In this paper we focus on predicting future instance semantic segmentations of moving objects, since we believe it is the most informative representation for a machine planning motion.
Most approaches are multimodal~\cite{graber2021panoptic}, integrating multiple modalities to achieve the highest accuracy. In general semantic segmentation~\cite{luc2017predicting}, optical flow~\cite{terwilliger2019recurrent} and deep features~\cite{luc2018predicting} are considered as a source for future prediction.
In this paper we study how can we rely on optical flow as unique source of information to forecast instance segmentations.
We break down the problem in two steps: optical flow forecasting and learned instance warping. Our contribution is twofold:
\begin{itemize}
\item We first design an optical flow predictor to generate future flows based on convolutional LSTM and a UNet architecture.
\item We train a warping neural network, with a specialized loss, that propagates current instance segmentations onto future frames, guided by predicted optical flows.
\end{itemize}

We show that, by choosing the appropriate loss and by learning a warp operator over noisy autoregressive inputs, we can obtain state-of-the art results on par with more complex multi-modal multi-scale approaches.

\section{Related Work}
Anticipating the future is a desirable asset for decision-making systems in variegate applications, like autonomous driving and robot navigation. In these scenarios, an artificial agent should have an intelligent component that forecasts motion of surrounding agents so to navigate safely.
Huge progress has been done on anticipating the future by addressing different tasks, also exploiting multiple modalities.
Early work addressed the video prediction task, where RGB past frames are used to synthesize future frames~\cite{oprea2020review}.
A large variety of methods was introduced, including autoregressive models~\cite{kalchbrenner2017video}, adversarial training~\cite{mathieu2015deep}, bidirectional flow~\cite{kwon2019predicting}, 3D convolutions~\cite{wang2018eidetic} and recurrent networks~\cite{villegas2017decomposing} to improve prediction accuracy. Instead of predicting raw RGB values, in this work we are interested in directly forecasting more meaningful semantic-level contexts for future frames. 
Several works have been recently proposed addressing semantic segmentation and instance segmentation forecasting.

\paragraph{\textbf{Future Semantic Segmentation}}
Scene understanding through future semantic segmentation can help an autonomous agent to take more intelligent decisions, e.g. forecasting specific category classes like pedestrians~\cite{brazil2017illuminating} or vehicles~\cite{chen2016monocular}.
Luc et al.~\cite{luc2017predicting} proposed to learn semantic-level scene dynamics by mapping semantic segmentations, which are extracted from past frames, to the future through a deep CNN. In this work the authors also shown that directly predicting semantic segmentation is more effective than predicting RGB frames and then fed them into a semantic segmentation model.
Terwilliger et al.~\cite{terwilliger2019recurrent} designed a flow-guided model, which first aggregates past flow features using a recurrent network to predict the future optical flow; then, the predicted flow is used to learn a warp layer to map the last semantic segmentation to the future next one.
Chiu et al.~\cite{chiu2020segmenting} proposed a teacher-student learning network, which encodes past RGB frames to directly generate future segmentations.
Other recent approaches extract features at different resolutions from past frames in order to map to the future ones, taking inspiration from the F2F architecture designed in~\cite{luc2018predicting}. \v{S}ari{\'c} et al.~\cite{vsaric2019single} used deformable convolutions to F2F network at a single-level in order to encode varied motion pattern. In \cite{saric2020warp}, the same authors enriched the features extracted from past images with correlation coefficients between neighbouring frames, and performed the forecasting by warping observed features according to regressed feature flow.

We propose to learn motion dynamics through an optical flow predictive model, using an encoder-decoder model with skip connections followed by a ConvLSTM~\cite{xingjian2015convolutional}. Differently from~\cite{terwilliger2019recurrent} and~\cite{saric2020warp}, we generate future semantic segmentations of moving objects, by aggregating together warped object masks, using the predicted flow and the semantic segmentation as inputs. Specifically, our model forecasts single instance segmentations, which allows a finer grained understanding of the scene for tasks such as path planning.
In addition, we exploit a fully learnable warping module instead of relying on a differentiable, yet not-learnable, warping operator as in~\cite{terwilliger2019recurrent}.

\paragraph{\textbf{Future Instance Segmentation}}
The goal of the future instance segmentation task is to detect and segment individual objects of interest appearing in a video sequence.
Semantic segmentation does not account for single objects, since they are fused together by assigning the same category label. Moreover instances can vary in number between frames and are not consistent across a video sequence.
Most recent methods generate future instance and semantic segmentations, by forecasting features for unobserved frames given the past ones. For instance, Luc et al.~\cite{luc2018predicting} designed a multi-scale approach, named F2F, based on a convolutional network with Feature Pyramid Networks (FPN)\cite{lin2017feature}. Then, future instance segmentations are detected by running Mask R-CNN\cite{he2017mask} from the predicted features, while future semantic segmentations, are computed by converting instance segmentation predictions according to the confidence score, thus proving better qualitative results than \cite{luc2017predicting}. Sun et al.~\cite{sun2019predicting}, following the same pipeline in~\cite{luc2018predicting}, employs a set of connected ConvLSTMs~\cite{xingjian2015convolutional}, so to capture rich spatio-temporal contexts across different pyramid levels.
Hu et al.~\cite{hu2021apanet} proposed a multi-stage optimization framework that exploits auto-path connections so to adaptively and selectively aggregate contexts from previous steps.

Lin et al.~\cite{lin2021predictive} designed a three-stage framework, able to forecast segmentation features from a decoder through predictive features outputted by an encoder, instead of directly predict pyramid segmentation features from input frames. Different than F2F~\cite{luc2018predicting}, future predictions are finally generated by jointly training the whole system using Mask R-CNN~\cite{he2017mask} and Semantic FPN~\cite{kirillov2019panoptic}, for semantic segmentations and instance segmentations respectively. Graber et al.~\cite{graber2021panoptic}, proposed a more complete framework to forecast the near future, by decomposing a dynamic scene into \textit{things} and \textit{stuff}, i.e. individual objects and background, with multiple training stages and also considering odometry anticipation due to camera motion.
In summary, recent works mainly address the future prediction task, both in terms of instances and semantics, following the typical pipeline based on  F2F, using recurrent networks and complex connections. The predicted features are then fed in input to different segmentation models, so to provide instances and semantic predictions. Although taking several training stages, such frameworks are able to get better results.
Our method, instead, proposes to combine the predicted dense optical flow with an instance warping network, without taking several optimization stages. We also prove that the warping model can learn to forecast object motion, from the predicted scene flow and the current semantic map.

\section{Our Approach}
\label{sec:method}
\begin{figure}[t]
\begin{tabular}{c}
\includegraphics[width=1.0\textwidth]{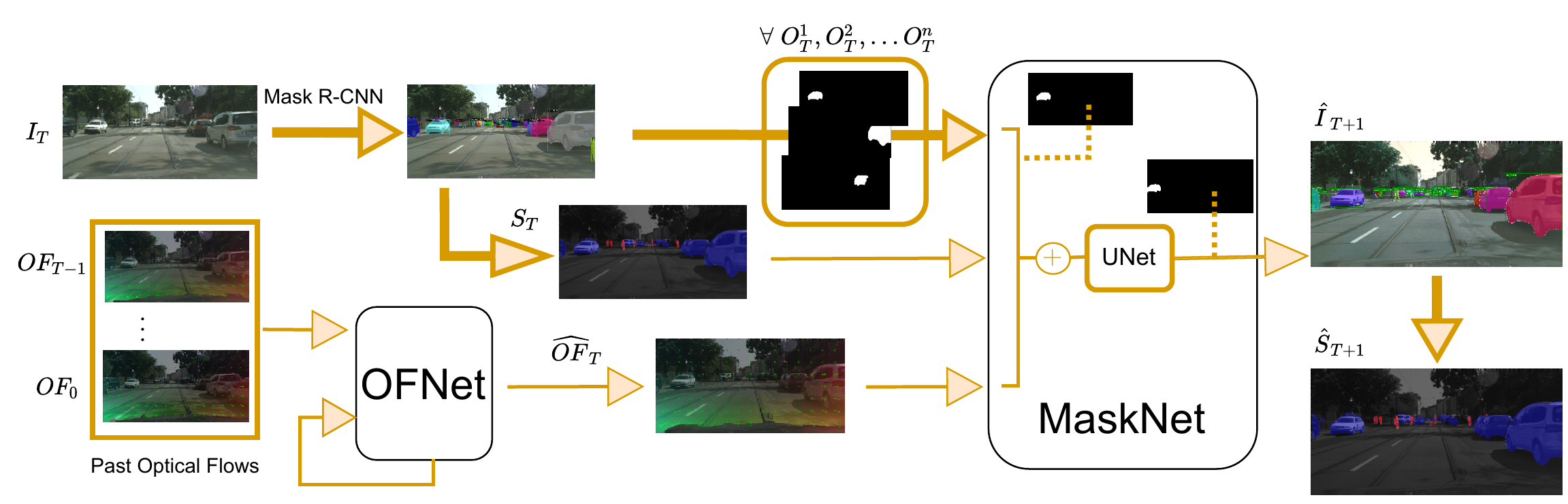}
\end{tabular}
\caption{OFNet forecasts future flows from past ones autoregressively. MaskNet warps instance segmentations to the future with the predicted flow. Future semantic segmentations are obtained by aggregating instances together.}
\label{fig:ourapproach}
\end{figure}

We introduce a novel framework for future instance segmentation that learns to predict future flow motion patterns. Such predictions are used along with the last observed semantic segmentation to warp detected objects to future unobserved frames (Fig.~\ref{fig:ourapproach}).
This architecture consists of two main components: (i) OFNet, an optical flow predictive model, which forecasts motion dynamics through past optical flows, and (ii) MaskNet, a novel mask prediction model, trained to warp segmentation instances of moving objects to future frames.

Warping is performed by leveraging a region-based loss function instead of the typical cross-entropy loss, so to deal with small parts of interest since the background occupies most of the scene.
To identify instances in a video sequence, we run Mask R-CNN~\cite{he2017mask} for each frame. Then, we exploit a bounding box association-based tracker based on~\cite{cuffaro2016segmentation} to generate object tracks and link instances for the sequence lifespan. Inspired by \cite{luc2018predicting}, the proposed approach is also able to generate semantic segmentations by fusing instance mask predictions according to a rescoring function balancing detector score and mask size.

\subsection{Optical Flow Forecasting}
\label{sec:flow}
The first key part of the framework is OFNet, the optical flow prediction model. It is a supervised network that learns to generate future flows based on past ones. OFNet takes $T$ dense optical flows from consecutive past frames and predicts the flow for the following time step.

\begin{figure}[t]
\centering
\begin{tabular}{ccc} 
\includegraphics[valign=t,width=.3\linewidth]{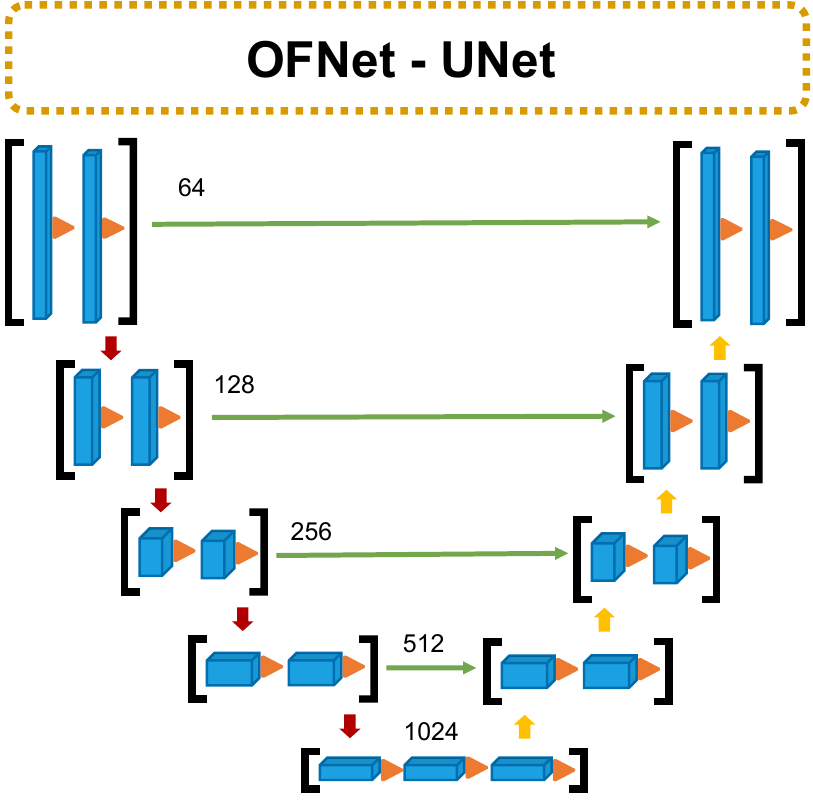} \quad \quad & \includegraphics[valign=t, width=.3\linewidth]{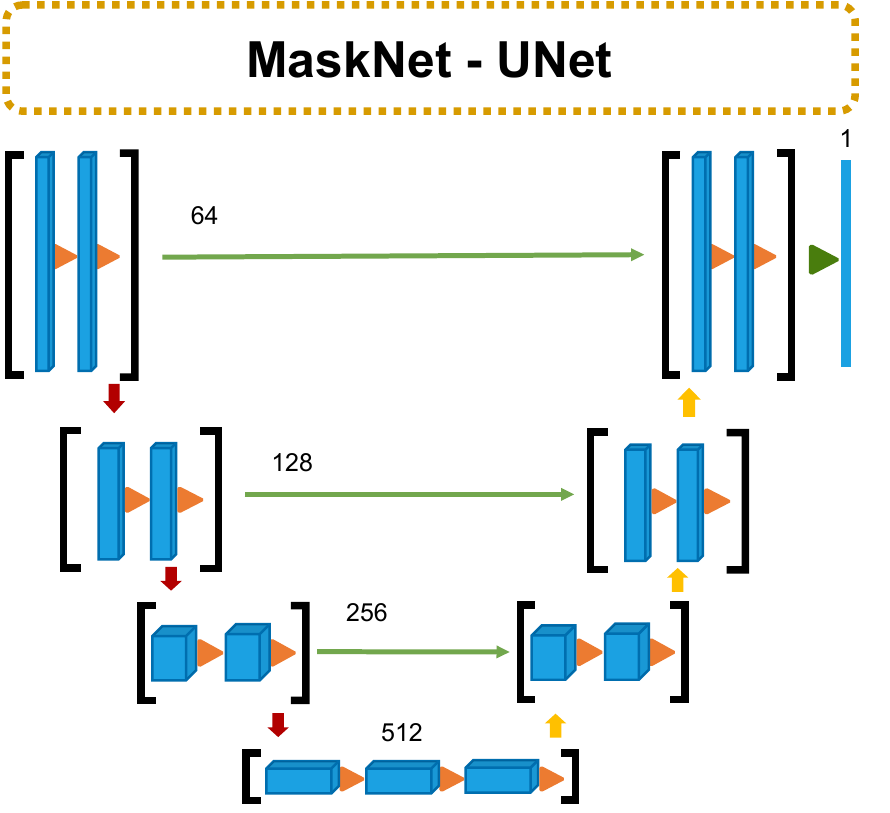} & \quad \includegraphics[valign=t, width=.15\linewidth]{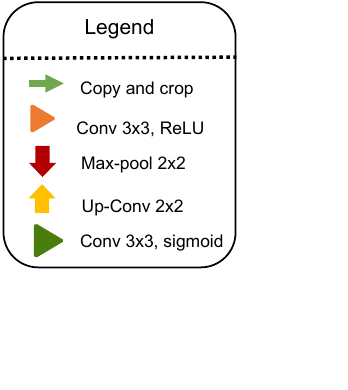}\\
\parbox[c][][c]{0.05\textwidth}{\subfigure[]{\label{fig:unetof}}} & \parbox[c][][c]{0.05\textwidth}{\subfigure[]{\label{fig:unetmasknet}}}
\end{tabular}
\caption{UNet architectures used as feature extractor \subref{fig:unetof} to generate 64-channel flow features in OFNet, and \subref{fig:unetmasknet} to produce future binary masks in MaskNet.}\label{fig:unet}
\end{figure}

Each optical flow is first estimated from consecutive pairs of frames, by running FlowNet2~\cite{ilg2017flownet}, that produces a dense 2D flow field, with $(u,v)$ horizontal and vertical displacements for each pixel.
Then, we learn motion features from each of these optical flows through a time distributed UNet~\cite{ronneberger2015u}, a fully convolutional encoder-decoder network with a contracting path and an expanding path, connected with skip connections. From the original UNet we consider the entire structure up to the final prediction layer, so to output features with 64 channels (Fig. \ref{fig:unetof}). The sequence of extracted features generated by UNet are fed to a ConvLSTM~\cite{xingjian2015convolutional} with $3 \times 3$ kernel followed by a $1 \times 1$ convolution, so to reduce the output channels to 2 as in a flow field.
We train OFNet to forecast an output sequence shifted by one step ahead with respect to input, i.e. by generating the next flow for each corresponding input. This provides supervision for each timestep and allows us to use it in an autoregressive fashion.

Formally, let $OF_i$ be the optical flow produced by FlowNet2 given two consecutive frames $X_i$ and $X_{i+1}$.
We denote with $F_i$ the motion features extracted from $OF_i$ using our UNet-based backbone. We extend such notation referring to $F_{t:T}$ as the sequence of motion features spanning from $t$ to $T$.
Likewise, we use the same notation for optical flow sequences $OF_{t:T}$.
At training time, given a sequence of $T$ flow features $F_{t:t+T}$, OFNet learns to generate a sequence of $T$ consecutive optical flows shifted by one timestep $\widehat{OF}_{t+1:t+T+1}$.
Instead, at inference time we only retain the last prediction of the output sequence $\widehat{OF}_{t+T+1}$, that encodes the scene motion towards the first unseen future frame.
OFNet is trained using an L2 loss function on the output sequence $\mathcal{L}_{\text{flow}}$:
\begin{equation}
\mathcal{L}_{\text{flow}}= \frac{1}{T} \mathlarger{\mathlarger{\sum}}_{k=1}^{T} \frac{\left(OF_k - \widehat{OF}_k\right)^2}{H \cdot W \cdot 2}
\label{eq:mse-lossfunction}
\end{equation}
where T is the sequence length and H and W are the height and width of the feature map. We set $T=6$ for all our experiments to provide sufficient information about past dynamics.
\subsection{Instance Mask Forecasting}
\label{sec:mask}
After generating future optical flows, we forecast future instance segmentations. To do that, we designed MaskNet, a novel model which leverages motions encoded in the predicted flows and the semantic segmentation of an observed frame, which provides context about the objects in the scene.
Instead of globally warping semantic segmentations as done in~\cite{terwilliger2019recurrent}, MaskNet is trained to understand future positions of individual objects. We aggregate information from the present frame $X_{t}$, i.e. the semantic segmentation of the scene $S_t$ and a binary mask of the instance to be warped $M^j_t$, and from the estimated future, i.e. the predicted flow field $\widehat{OF}_t$. The data is concatenated channel-wise and fed to the network, which generates the future instance mask $\widehat{M}^j_{t+1}$ located in frame $X_{t+1}$.

As in OFNet, we use a UNet backbone for MaskNet but we reduce its encoder-decoder structure by one level to avoid overfitting (Fig. \ref{fig:unetmasknet}).
After aggregating $\{\widehat{OF}_t, S_t, M^j_t\}$, MaskNet takes as input a tensor of size  $H \times W \times 4$ and produces as output a binary map $H \times W \times 1$, where 0 corresponds to background and 1 to the forecasted instance mask. To binarize the output we use a sigmoid activation and we threshold at 0.5.
Since an object mostly occupies a very small region compared to the input resolution, we are in the typical case of class unbalance. To address this issue, rather than training MaskNet with a cross-entropy loss, we use a Dice loss~\cite{milletari2016v}. Dice loss is a region-based loss that helps to better focus on overlapping regions in two images, which in our case are the predicted binary mask and the corresponding groundtruth instance.
Thus, considering an $N$ pixel image, we employ the mask loss $\mathcal{L}_{\text{mask}}$ defined through the Dice coefficient $D$:
\begin{equation}
\mathcal{L}_{\text{mask}} = 1 - D = 1 - \frac{2 \sum_{i=1}^{N} \hat{M}_i \; {M^{GT}}_i}{\sum_{i=1}^N \hat{M}^2_i + \sum_{i=1}^N {M^{GT}}^2_i}
\label{eq:dice-lossfunction}
\end{equation}
where the Dice coefficient $D\in[0,1]$ measures the similarity between the predicted mask $\hat{M}$ and the groundtruth one $M^{GT}$. Note that to compute the loss, we use the output of MaskNet after the sigmoid activation and before the final thresholding in order to work with real values in $[0,1]$.
Since our model works with single instances, we do not rely on a dedicated network to generate semantic segmentations of the image as most state of the art methods~\cite{graber2021panoptic}\cite{terwilliger2019recurrent}\cite{lin2021predictive}. Instead, we simply group all object masks together maintaining their category label.

\section{Experiments}
We conduct experiments on the Cityscapes dataset~\cite{cordts2016cityscapes}, which contains recordings from a car driving in urban environments for a total of 5000 sequences, divided in 2975 for train, 1525 for testing and 500 for validation. Every sequence is long 1.8s and consists of 30 frames at resolution $1024 \times 2048$. Groundtruth semantic and instance segmentations are available for the 20-th frame of each sequence.
Our method is tested on 8 moving object categories: \emph{person}, \emph{rider}, \emph{car}, \emph{truck}, \emph{bus}, \emph{train}, \emph{motorcycle}, \emph{bicycle}.
We evaluate the future predictions, both instance and semantic segmentations, 3 and 9 frames ahead, respectively denoted as short-term \textcolor{black}{(about 0.17s later)} and mid-term \textcolor{black}{(about 0.5s later)}. Such evaluations are performed on annotated frames in the validation set, as done in literature.
Predicted optical flows are obtained, by initially feeding $T=6$ consecutive past optical flows $OF_{t-6:t-1}$ to OFNet and then, autoregressively forecasting each future flow $OF_{t+k}$ with $k=0, 1, \ldots, n$ ($n=2$ for short-term and $n=8$ for mid-term).
Future semantic segmentation is evaluated by measuring the intersection-over-union (IoU) between predictions and the corresponding ground truth for each class and averaging among classes. Future instance segmentation instead is evaluated through two standard Cityscapes metrics: AP50 and AP. AP50 is average precision where an instance is correctly counted if it has at least 50\% of IoU with the groundtruth. AP is average precision, averaged over ten equally spaced IoU thresholds from 50\% to 95\%.
Since these metrics require a confidence score for each prediction, we use the score provided by the detector for each object.
However, small objects are likely to be false positives generated by the detector. To mitigate this issue, we reduce the score by 0.5 for objects smaller than 64x64px and by 0.3 for objects smaller than 128x128px.

\begin{table}[t]
\caption{Effect of different training regimes for MaskNet. Warping instances with predicted flows is included as a baseline.
}\label{table:ablation}
\begin{adjustbox}{width=\textwidth}
\begin{tabular}[t]{c|cc|c|cccccc}
\toprule
\multirow{2}{*}{Method \quad} & \multicolumn{2}{c|}{Train Data} & Finetuned & \multicolumn{3}{c}{Short term $\Delta t=3$} & \multicolumn{3}{c}{Mid term $\Delta t=9$}\\
& \quad FlowNet \quad & \quad OFNet \quad \quad & Layers & \quad AP \quad & \quad AP50 \quad & \quad IoU \quad & \quad AP \quad  & \quad AP50 \quad & \quad IoU \quad\\ \midrule
Warping & \xmark & \xmark & - & 15.9 & 34.3 & 60.7 & 2.8 & 8.8 & 39.4\\\hline
\multirow{4}{*}{MaskNet} & \cmark & \xmark & - & 18.7 & 39.4 & 65.5 & 4.0 & 12.3 & 44.5\\

& \xmark & \cmark & all & 18.5 & 39.3 & 65.2 & 4.5 & 14.9 & 41.9\\
& \cmark & \cmark & 3 & 18.7 & 39.2 & 65.7 & 5.1 & 16.0 & 44.8 \\
& \cmark & \cmark & 2 & \textbf{18.9} & \textbf{39.5} & \textbf{65.9} & \textbf{5.9} & \textbf{17.4} & \textbf{45.5}\\
\bottomrule
\end{tabular}
\end{adjustbox}
\end{table}

\subsection{Implementation Details and Ablation Study}
We compute the optical flow at every pair of consecutive frames for each video sequence, both for training and validation sets, through FlowNet-c with pretrained weights from FlowNet2~\cite{ilg2017flownet}. Frames are resized to $128 \times 256$, as in \cite{terwilliger2019recurrent}.
We detect all objects from each video frame, by running the Mask R-CNN model provided in Detectron2~\cite{wu2019detectron2}, pretrained on COCO~\cite{lin2014microsoft} and finetuned on Cityscapes.
We train OFNet with Adam and learning rate 0.0001. Our architecture is inspired by UNet,  but in order to perform regression we change  the number of channels to 2 in the last $1 \times 1$ convolutional layer and we use a linear activation function.  We set the batch size to 3 and we employ the L2 loss as in Eq.~\ref{eq:mse-lossfunction} and we train the network for 100 epochs.
MaskNet instead is trained for 4 epochs using Adam with learning rate $0.0001$. Inputs are resized to $256 \times 512$. First, we pretrain the model using optical flows computed with FlowNet~\cite{ilg2017flownet} instead of the predicted ones. Then, we finetune for 3 additional epochs the last two layers using the predicted flows autoregressively, for both short-term and mid-term models. This two-step optimization allows the model to learn how to warp instances using clean flows and then to adapt itself to handle noisy flow fields.

We observe that using a Cross-Entropy loss instead of a Dice loss, MaskNet converges to trivial solutions predicting background for every pixel of the frame. This underlines the importance of using a region based loss for this kind of task.

In Tab.~\ref{table:ablation} we conduct an ablation study by exploring different training strategies. As a first control experiment, we output future segmentations by directly warping masks with flows predicted by OFNet. In this case MaskNet is not used. Then we evaluate the effect of pretraining MaskNet using ground truth flows generated by FlowNet~\cite{ilg2017flownet}. Different fine-tuning strategies are also analyzed.

As expected using a learned warping function, MaskNet, yields superior results with respect to simply warping masks.
It also emerges that the two step training is fundamental. Training MaskNet only on precomputed flows yields good results at short term but exhibits a severe drop in performance at mid-term, since the model is not trained to deal with noisy predictions as inputs. Similarly, using only predicted flows, i.e. without pretraining, the model fails to warp masks appropriately up to mid-term.
We also study the effect of using OFNet in an autoregressive way or by feeding FlowNet flows to it in a teacher-forcing way. This allows us to better understand the impact of flow quality on MaskNet's ability to warp masks. Interestingly, when OFNet is fed autoregressively with predicted flows, MaskNet exhibits a large improvement in average precision for mid-term predictions. In fact, using a teacher forcing approach, i.e. feeding only clean flows to OFNet, the produced flows are not sufficiently realistic. This is confirmed by the fact that this model performs similarly to the one trained only on real flows.
Finally, we compare performances by finetuning the last 3 layers instead of 2. We found that 2 layers to train is the best choice.

\begin{table}[t]
\caption{Flow prediction MSE up to mid-term (t+9) on Cityscapes val set. We also show a break down of MSE for the $u$ and $v$ optical flow components.}
\centering
\begin{tabular}{l | ccccccccc}
\toprule
 & ~~t+1~~ & ~~t+2~~ & ~~t+3~~ & ~~t+4~~ & ~~t+5~~ & ~~t+6~~ & ~~t+7~~ & ~~t+8~~ & ~~t+9~~ \\ \hline
$MSE$ & 0.36 & 0.54 & 0.60 & 0.78 & 0.84 & 0.99 & 1.08 & 1.28 & 1.19 \\
$MSE_u$ & 0.45 & 0.67 & 0.70 & 0.94 & 1.04 & 1.24 & 1.31 & 1.64 & 1.55 \\
$MSE_v$ & 0.26 & 0.41 & 0.50 & 0.63 & 0.63 & 0.75 & 0.84 & 0.91 & 0.82 \\
\bottomrule
\end{tabular}
\label{tab:mse}
\end{table}

\section{Results}
\label{sec:results}
We first analyse the prediction capabilities of OFNet. Since we want to use it to generate future flows up to mid-term (9 frames in the future), we show in Tab.~\ref{tab:mse} how error accumulates while predicting flows autoregressively. As expected, the MSE values increase as the time horizon shifts forward, although the error accumulation appears to be bounded by a linear growth. Interestingly, the horizontal component $u$ exhibits a higher error. This is caused by the fact that most objects move horizontally in front of the camera capturing an urban scene from an ego-vehicle perspective.

\begin{figure}[t]
\begin{minipage}[t]{0.4\textwidth}
\centering
\begin{adjustbox}{width=\textwidth}
\begin{tabular}{ccc}
& & \\
\makecell{\textcolor{green}{Input}\\vs\\\textcolor{red}{Output}} & \makecell{\textcolor{green}{Input}\\vs\\\textcolor{red}{Pred}} & \makecell{\textcolor{green}{Output}\\vs\\\textcolor{red}{Pred}}\\
\includegraphics[trim={120px 35px 100px 35px},clip,valign=m,width=.3\textwidth]{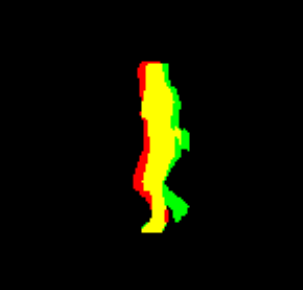} & \includegraphics[trim={120px 35px 100px 35px},clip,valign=m,width=.3\textwidth]{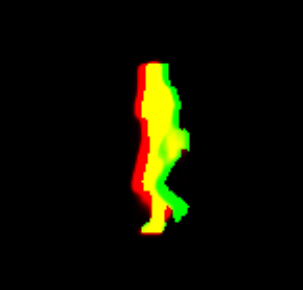} & \includegraphics[trim={120px 35px 100px 35px},clip,valign=m,width=.3\textwidth]{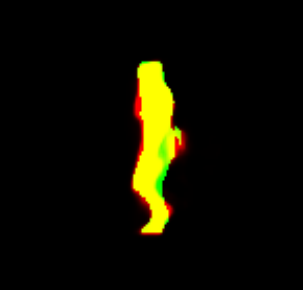}\\ \midrule
\includegraphics[trim={40 35px 180px 35px},clip,valign=m,width=.3\textwidth]{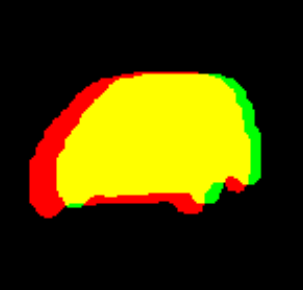} & 
\includegraphics[trim={40 35px 180px 35px},clip,valign=m,width=.3\textwidth]{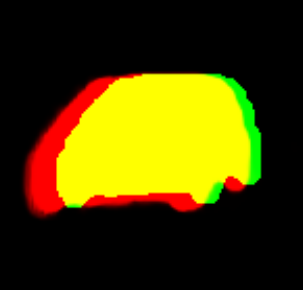} & \includegraphics[trim={40 35px 180px 35px},clip,valign=m,width=.3\textwidth]{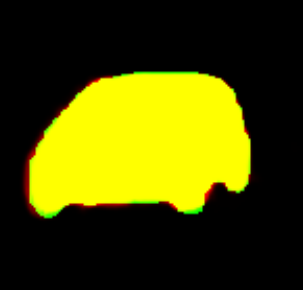}\\ \midrule
\includegraphics[trim={55 35px 165px 35px},clip,valign=m,width=.3\textwidth]{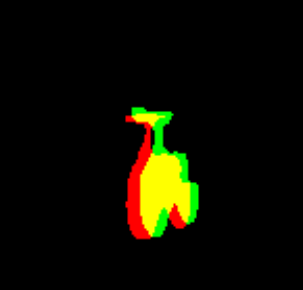} & \includegraphics[trim={55 35px 165px 35px},clip,valign=m,width=.3\textwidth]{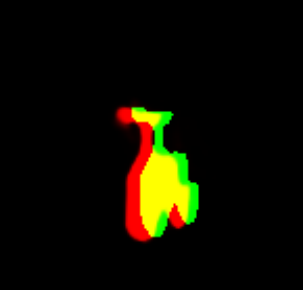} & \includegraphics[trim={55 35px 165px 35px},clip,valign=m,width=.3\textwidth]{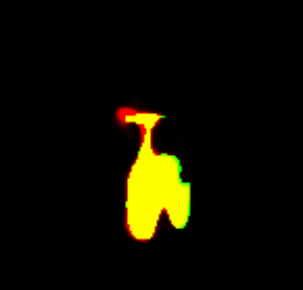}\\
\end{tabular}
\end{adjustbox}
\captionof{figure}{Instance predictions: input vs gt output, input vs prediction, gt output vs prediction. Mask intersections in yellow.}\label{fig:warpbyopticalflow}
\end{minipage}
\quad
\begin{minipage}[t]{0.58\textwidth}
\centering
\begin{adjustbox}{width=\textwidth}
\begin{tabular}{@{}ccccccc@{}}
\toprule
Method & \multicolumn{3}{c}{Short term $\Delta t=3$} \quad & \multicolumn{3}{c}{Mid term $\Delta t=9$} \\ 
 & AP & AP50 & IoU & AP & AP50 & IoU \\ \midrule
Mask RCNN oracle & 34.6 & 57.4 & 73.8 & 34.6 & 57.4 & 73.8 \\ \midrule
MaskNet-Oracle & 24.8 & 47.2 & 69.6 & 16.5 & 35.2 & 61.4\\ \midrule
Copy-last segm. \cite{luc2018predicting} & 10.1 & 24.1 & 45.7 & 1.8 & 6.6 & 29.1 \\
Optical-flow shift \cite{luc2018predicting} & 16.0 & 37.0 & 56.7 & 2.9 & 9.7 & 36.7 \\
Optical-flow warp \cite{luc2018predicting} & 16.5 & 36.8 & 58.8 & 4.1 & 11.1 & 41.4 \\
Mask H2F\cite{luc2018predicting} & 11.8 & 25.5 & 46.2 & 5.1 & 14.2 & 30.5 \\
S2S\cite{luc2017predicting} & - & - & 55.4 & - & - & 42.4 \\
F2F\cite{luc2018predicting} & 19.4 & 39.9 & 61.2 & 7.7 & 19.4 & 41.2 \\
LSTM M2M\cite{terwilliger2019recurrent} & - & - & 65.1 & - & - & 46.3 \\
Deform F2F\cite{vsaric2019single} & - & - & 63.8 & - & - & 49.9 \\
CPConvLSTM\cite{sun2019predicting} & 22.1 & 44.3 & - & 11.2 & 25.6 & - \\
F2MF\cite{saric2020warp} & - & - & 67.7 & - & - & 54.6 \\
PSF\cite{graber2021panoptic} & 17.8 & 38.4 & 60.8 & 10.0 & 22.3 & 52.1 \\
APANet\cite{hu2021apanet} & 23.2 & 46.1 & 64.9 & 12.9 & 29.2 & 51.4 \\
PFL\cite{lin2021predictive} & 24.9 & 48.7 & 69.2 & 14.8 & 30.5 & 56.7 \\ \midrule
MaskNet (Ours) & 19.5 & 40.5 & 65.9 & 6.4 & 18.4 & 45.5 \\
\bottomrule
\end{tabular}
\end{adjustbox}
\captionof{table}{Comparison results for future instance segmentation (AP and AP50) and future semantic segmentation (IoU) of moving objects on Cityscapes val set.}\label{table:results}
\end{minipage}
\end{figure}

We then provide in Fig.~\ref{fig:warpbyopticalflow} a qualitative analysis of the warping capabilities of MaskNet feeding clean future flows using FlowNet2 as an oracle. In the following we will refer to this model as \textit{MaskNet-Oracle}.

\begin{figure}[t]
\begin{adjustbox}{width=\textwidth}
\begin{tabular}{lcc|cc}
& \multicolumn{2}{c}{Short-term $\Delta t=3$} & \multicolumn{2}{c}{Mid-term $\Delta t=9$} \\
& Mask R-CNN Oracle & Our predictions & Mask R-CNN Oracle & Our predictions \\
\multirow{2}{*}{\makecell{Instance \\ Segmentation}} & \includegraphics[width=.25\linewidth]{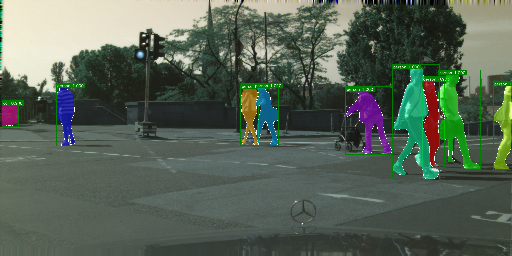} & \includegraphics[width=.25\linewidth]{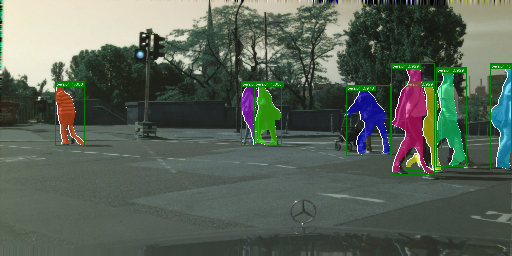} &
\includegraphics[width=.25\linewidth]{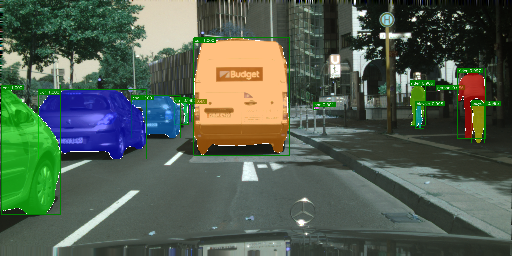} &  \includegraphics[width=.25\linewidth]{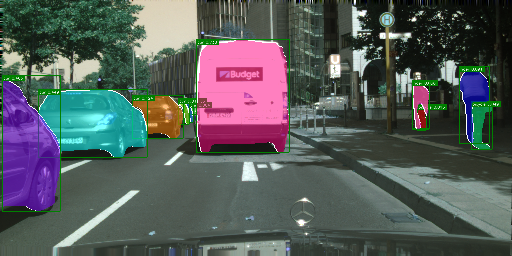}\\

& \includegraphics[width=.25\linewidth]{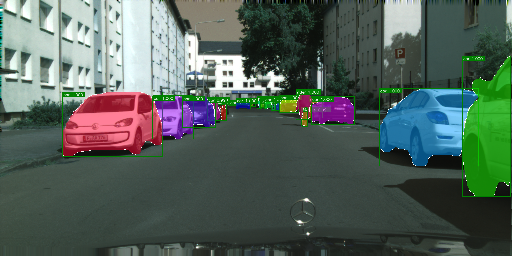} & \includegraphics[width=.25\linewidth]{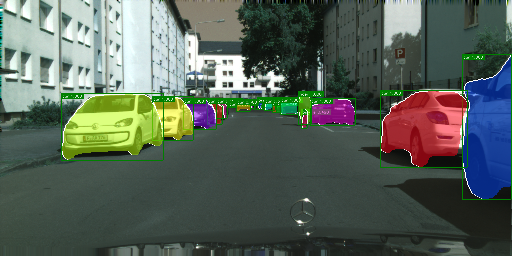} &
\includegraphics[width=.25\linewidth]{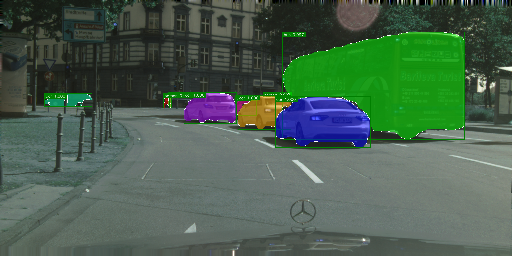} & \includegraphics[width=.25\linewidth]{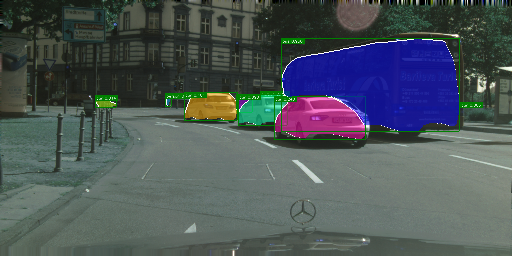}\\

\multirow{2}{*}{\makecell{Semantic \\ Segmentation}} & \includegraphics[width=.25\linewidth]{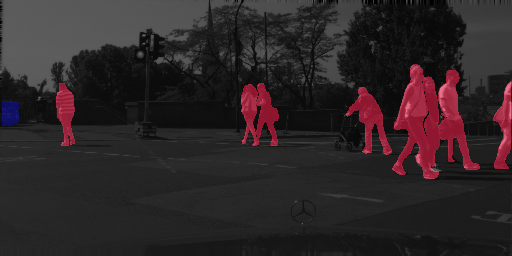}  & \includegraphics[width=.25\linewidth]{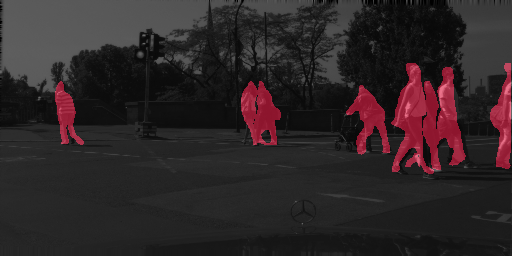} &
\includegraphics[width=.25\linewidth]{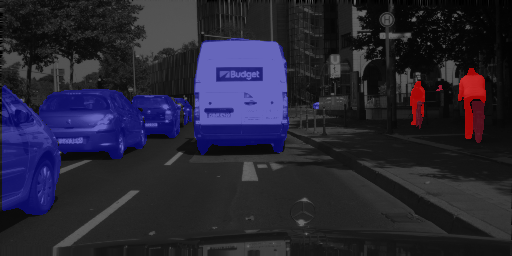} &  \includegraphics[width=.25\linewidth]{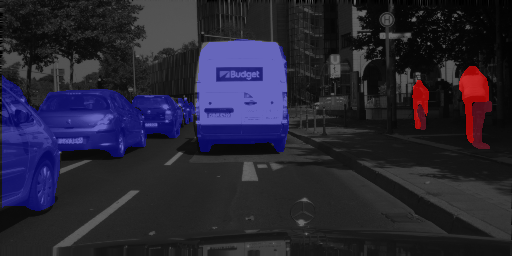}\\

& \includegraphics[width=.25\linewidth]{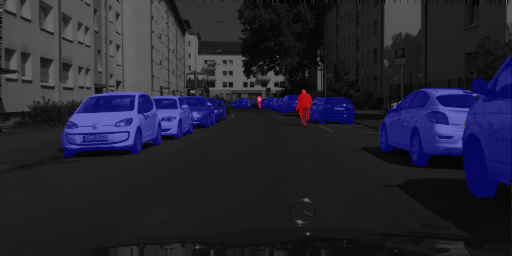} & \includegraphics[width=.25\linewidth]{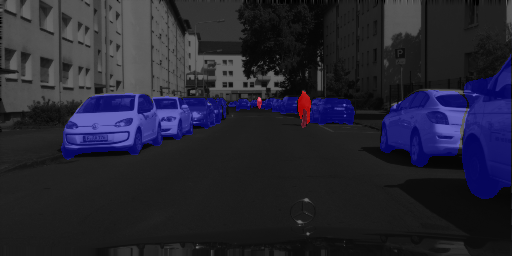} &
\includegraphics[width=.25\linewidth]{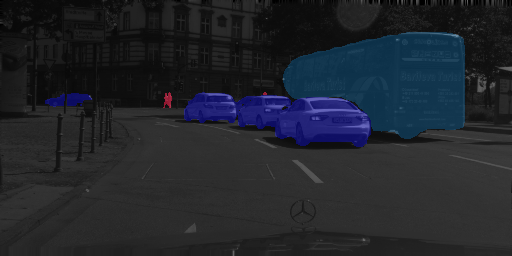} &  \includegraphics[width=.25\linewidth]{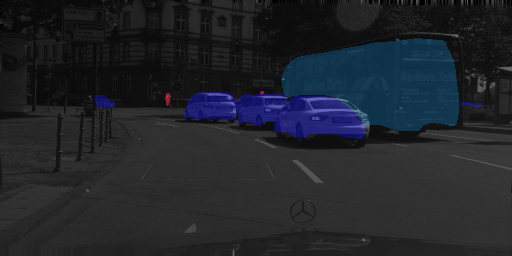} \\
\end{tabular}
\end{adjustbox}
\caption{Visualization results of future predictions on Cityscapes val set.}\label{fig:qualitativeresults}
\end{figure}

This model can provide accurate results among different categories. Interestingly, it can correctly warp small non-rigid parts of an instance, like legs, as well as objects at different scales like approaching cars.
These results prove that MaskNet can understand most motions of individual objects in a scene by simply exploiting optical flow, without further motion information.
MaskNet-Oracle, being fed with ground truth optical flows, acts as an upper bound of our system. We also evaluate a Mask R-CNN oracle, by directly running Mask R-CNN on future frames.
We compare our approach, that uses the predicted optical flows from OFNet, with such upper bounds and with recent approaches from the state of the art (Tab.~\ref{table:results}).
We exceed some state of the methods, such as F2F and PSF~\cite{graber2021panoptic} at short-term, while mid-term predictions are very close to the feature-to-feature approach introduced by Luc et al.~\cite{luc2018predicting}. This confirms our choice of using motion cues, instead of learning different Feature Pyramid Networks.
However, recent frameworks, such as CPConvLSTM~\cite{sun2019predicting}, APANet~\cite{hu2021apanet} and PFL~\cite{lin2021predictive}, reported better results, where more complex information from intra-level and inter-level connections are learned across FPNs from past frames.
This requires several training stages to improve both short and mid predictions.
We also generate future semantic segmentations for moving objects (Tab.~\ref{table:results}). We obtain the future semantic segmentations by grouping by category the predicted instance masks.
MaskNet obtains good results and overcomes most of those state-of-the-art approaches that provide forecasts in terms of both instance and semantic segmentations.
In general our approach performs well at short-term, but gets worse at mid-term compared to methods that predict future FPNs after several training stages and learning specific connections, like the auto-path connections as in APANet~\cite{hu2021apanet}, or that use segmentation models properly trained on predicted features, as done in PFL~\cite{lin2021predictive}.
Some qualitative results are shown for instance and semantic segmentation predictions, both at short-term and mid-term, in Fig.~\ref{fig:qualitativeresults}.

Overall, compared to the other methods, our approach is simple and without complex optimization stages can reach competing performances. Nonetheless, our approach underlines the efficacy and the limitations of relying solely on optical flow data, which is easy to obtain even with any typical on-board sensor whether it is as RGB camera, an event camera or a LiDAR.
However, we believe that there is still margin for improvement using a flow-based predictor like ours.
This stems from the results of our MaskNet-Oracle, as using clean flows estimated by FlowNet2 leads to overcome all state-of-the-art results at both short-term and mid-term.
We believe that integrating information such as depth-maps could lead to significant improvements in the state of the art. However, this is out of the scope of our paper and we leave it as future work.

\section{Conclusions}
In this work we addressed the future instance segmentation task using predicted optical flows and a learned warping operator, which also allows to produce future semantic segmentations without additional training. We designed a framework made of a predictive optical flow model able to forecast future flows given past ones through a UNet+ConvLSTM architecture. A second component, MaskNet, warps observed objects to future frames, using the current semantic map and the predicted optical flow.
We conducted experiments on Cityscapes, demonstrating the effectiveness of optical flow based methods.

\begin{scriptsize}
\paragraph{\textbf{Acknowledgements}}
This work was supported by the European Commission under European Horizon 2020 Programme, grant number 951911 - AI4Media
\end{scriptsize}

%
%
\bibliographystyle{splncs04}
\bibliography{ref.bib}
\end{document}